\documentclass{article} 
\usepackage{iclr2024_conference,times}

\usepackage{amsmath,amsfonts,bm}

\def\eqref#1{equation~\ref{#1}}

\def\1{\bm{1}}

\DeclareMathAlphabet{\mathsfit}{\encodingdefault}{\sfdefault}{m}{sl}
\SetMathAlphabet{\mathsfit}{bold}{\encodingdefault}{\sfdefault}{bx}{n}

\usepackage{hyperref}
\usepackage{url}
\usepackage[percent]{overpic}

\usepackage{graphicx}
\usepackage{subcaption}
\usepackage{booktabs}
\usepackage{bbding}
\usepackage{xcolor}

\definecolor{codebrown}{rgb}{0.37,0.0,0.0}
\hypersetup{colorlinks,linkcolor={red},citecolor={codebrown},urlcolor={blue}}

\title{Consistent4D: Consistent 360\textdegree{} Dynamic Object Generation from Monocular Video}

\author{Yanqin Jiang$^1$, Li Zhang$^{3}$, Jin Gao$^1$, Weimin Hu$^1$, Yao Yao$^{2}$\,\textsuperscript{\Envelope} \\
$^1$CASIA, $^2$Nanjin University, $^3$Fudan University \\
jiangyanqin2021@ia.ac.cn, lizhangfd@fudan.edu.cn, \\
\{jin.gao, wmhu\}@nlpr.ia.ac.cn, yaoyao@nju.edu.cn
}

\iclrfinalcopy 
\begin{document}

\maketitle
\begin{abstract}

In this paper, we present Consistent4D, a novel approach for generating 4D dynamic objects from uncalibrated monocular videos.
Uniquely, we cast the 360-degree dynamic object reconstruction as a 4D generation problem, eliminating the need for tedious multi-view data collection and camera calibration.
This is achieved by leveraging the object-level 3D-aware image diffusion model as the primary supervision signal for training Dynamic Neural Radiance Fields (DyNeRF).
Specifically, we propose a Cascade DyNeRF to facilitate stable convergence and temporal continuity under the supervision signal which is discrete along the time axis.
To achieve spatial and temporal consistency, we further introduce an Interpolation-driven Consistency Loss.
It is optimized by minimizing the discrepancy between rendered frames from DyNeRF and interpolated frames from a pre-trained video interpolation model.
Extensive experiments show that our Consistent4D can perform competitively to prior art alternatives, opening up new possibilities for 4D dynamic object generation from monocular videos, whilst also demonstrating advantage for conventional text-to-3D generation tasks. Our project page is \href{https://consistent4d.github.io/}{https://consistent4d.github.io/}.

\end{abstract}

\begin{figure}[h]
    \centering
    \vspace{-5mm}
    \includegraphics[width=0.85\textwidth]{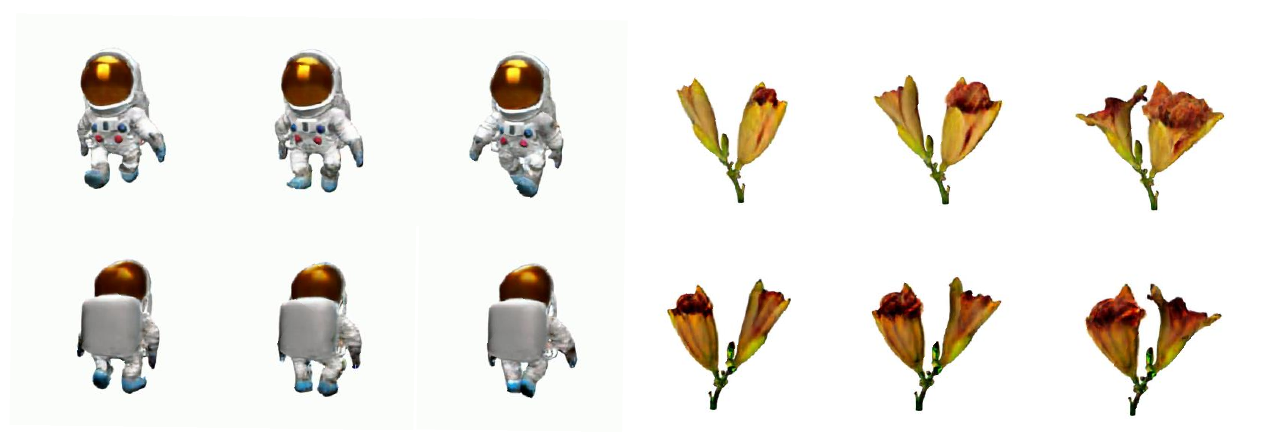}
    \caption{Video-to-4D results achieved by our method. We show the renderings of 2 objects at 2 viewpoints and 3 timestamps.}
    \label{fig:teaser}
\end{figure}
\section{Introduction}
Perceiving dynamic 3D information from visual observations is one of the fundamental yet challenging problems in computer vision, which is the key to a broad range of downstream applications e.g., virtual content creation, autonomous driving simulation, and medical image analysis. However, due to the high complexity nature of dynamic 3D signals, it is rather difficult to recover such information from a single monocular video observation. 
As a result, existing dynamic object \texttt{reconstruction} approaches usually take synchronized multi-view videos as inputs~\citep{li2022neural, li2022tava, shao2023tensor4d}, or rely on training data containing effective multi-view cues (e.g., teleporting cameras or quasi-static scenes~\citep{li2020neural, pumarola2021d, park2021hypernerf, park2021nerfies}). However, current reconstruction approaches often fail in reconstructing regions that were not observed in input sequences~\citep{gao2022monocular}. Moreover, multi-view data capturing requires synchronized camera rigs and meticulous calibrations, which inevitably limit the methods to potential real-world applications.

On the other hand, given only a video clip of a dynamic object, humans are capable of depicting the appearance, geometry, and movement of the object. This is achieved by prior knowledge of visual appearance accumulated through human life. In contrast to the multi-view video setting, our approach favors such simple and practical input settings of a static monocular video. The static monocular video offers several advantages: ease of collection for handheld cameras, minimizing the risk of motion blur due to camera movement, and obviating the need for camera parameter estimation. As a static monocular video does not provide effective multi-view information for reconstruction, we instead opt for the \texttt{generation} approaches which demonstrates diminished reliance on multi-view information.
 
In this work, we present a novel video-to-4D generation approach termed Consistent4D. 
In this approach, we represent dynamic objects through a specially designed Cascade DyNeRF and leverage a pre-trained 2D diffusion model to regulate the DyNeRF optimization, inspired by recent advancements in text-to-3D~\citep{poole2022dreamfusion, wang2022score, chen2023fantasia3d, wang2023prolificdreamer} and image-to-3D~\citep{deng2022nerdi, tang2023make, melas2023realfusion} techniques. 
The challenge is to achieve both spatial and temporal consistency.
To tackle this challenge, we introduce an Interpolation-driven Consistency Loss (ICL), which leverages a pre-trained video interpolation model to provide spatiotemporally coherent supervision signals.
Notably, the ICL loss not only enhances consistency in 4D generation but also mitigates multi-face issues in 3D generation. Furthermore, we train a lightweight video enhancer to enhance the video generated from dynamic NeRF as a post-processing step.

We have extensively evaluated our approach on both synthetic videos rendered from animated 3D model and in-the-wild videos collected from the Internet. To summarize, the contribution of this work includes:
\begin{itemize}
\item 
We propose a video-to-4D framework for dynamic object generation from a statically captured monocular video. A specially designed Cascade DyNeRF is applied to represent the object, optimized through the Score Distillation Sampling (SDS) loss by a pre-trained 2D diffusion model. Moreover, for the comprehensiveness of the framework, we train a video enhancer to improve the rendering of 4D object as a post-processing step.
\item 
To address the challenge of maintaining temporal and spatial consistency in 4D generation task, we introduce a novel Interpolation-driven Consistency Loss (ICL). The proposed ICL loss can significantly improve consistency, e.g., multi-face problem, in both video-to-4D and text-to-3D generation tasks. 
\item 
We extensively evaluate our method on both synthetic and in-the-wild videos collected from the Internet, showing promising results for the new task of video-to-4D generation.
\end{itemize}

\section{Related work}

\paragraph{3D Generation}
3D generation aims to generate 3D content conditionally or unconditionally.
Early works mainly make use of GAN, i.e., generate category-specific 3D objects or scenes from random noise after learning the category prior~\citep{schwarz2020graf, chan2021pi, gao2022get3d}.
Recently, general-purpose 3D generation has been enabled by text-to-image diffusion model pre-trained on Internet-scale data, and it also becomes more controllable, e.g., controlled by text prompt or single image.
The pioneer work of text-to-3D is DreamFusion~\citep{poole2022dreamfusion}, which proposes Score Distillation Sampling (SDS) loss to leverage the image diffusion model for neural radiance field (NeRF) training. 
The following works ~\citep{lin2023magic3d, chen2023fantasia3d, wang2023prolificdreamer} further enhance the visual quality of the generated object by using mesh representation, Variational Score Distillation, etc. 
However, the challenging problem, multi-face Janus problem, has not been addressed in the above works.
Image is another popular condition for 3D generation.
Different from 3D reconstruction, which focuses on the reconstruction of visible regions from multi-view images, 3D generation usually has only a single image and relies much on the image diffusion model to generate invisible regions of the object~\citep{melas2023realfusion, tang2023make, liu2023zero}.
So many works simply translate the image to words using Lora and then exploit text-to-3D methods~\citep{melas2023realfusion, tang2023make, seo2023let}.
One exception is Zero123~\citep{liu2023zero}, which trains a 3D-aware image-to-image model using multi-view data and could generate a novel view of the object in the input image directly.
Benefiting from multi-view data training, the multi-face Janus problem has been alleviated to some extent in Zero123.

\paragraph{4D Reconstruction}
4D reconstruction, aka dynamic scene reconstruction, is a challenging task.
Some early works focus on object-level reconstruction and adopt parametric
shape models~\citep{loper2023smpl, vo2020spatiotemporal} as representation.
In recent years, dynamic neural radiance field become popular, and convenient dynamic scene reconstruction is enabled.
These works can
be classified into two categories: a deformed scene is directly modeled as a NeRF in canonical space with a time-dependent deformation~\citep{pumarola2021d, park2021nerfies, park2021hypernerf, wu2022d, tretschk2021non} or time-varying NeRF in the world space~\citep{gao2021dynamic, li2020neural, xian2021space, fridovich2023k, cao2023hexplane}. 
Some of them require multi-view synchronized data to reconstruct dynamic scenes, however, data collection and calibration is not convenient~\citep{li2022neural, shao2023tensor4d}. 
So, reconstruction from monocular videos gain attention.
However, those monocular methods either require teleporting camera or quaic-static scenes~\citep{pumarola2021d, park2021nerfies, park2021hypernerf}, which are not representative of daily life scenarios~\citep{gao2022monocular}.

\paragraph{4D Generation}
4D generation extends 3D generation to space+time domain and thus is more challenging. 
Early works are mainly category-specific and adopt parametric shape models~\citep{zuffi20173d, Zuffi_2018_CVPR, vo2020spatiotemporal, kocabas2020vibe} as representation.
They usually take images or videos as conditions and need category-specific 3D templates or per-category training from a collection of images or videos~\citep{ren2021class, wu2021dove, yang2022banmo, wu2022magicpony}.
Recently, one zero-shot category-agnostic work, text-to-4D~\citep{singer2023text}, achieves general-purpose dynamic scene generation from text prompt. 
It follows DreamFusion~\citep{poole2022dreamfusion} and extends it to the time domain by proposing a three-stage training framework.
However, the quality of generated scenes is limited due to low-quality video diffusion models.

\section{Preliminaries}
\subsection{Score Distillation Sampling for image-to-3D}
Score Distillation Sampling (SDS) is first proposed in DreamFusion~\citep{poole2022dreamfusion} for text-to-3D tasks. It enables the use of a 2D text-to-image diffusion model as a prior for optimization of a NeRF.
We denote the NeRF parameters as $\theta$, text-to-image diffusion model as $\phi$, text prompt as $\rho$, the rendering image and the reimage as $\mathbf{x}$ and $\mathbf{z}$, the SDS loss is defined as:
\begin{equation}\label{eq: sds}
   \nabla_{\theta}\mathcal{L}_{SDS}\left(\phi, \mathbf{x}\right) =  \mathbb{E}_{\tau, \epsilon} \left[\omega(t) \left( \hat{\epsilon}_{\theta}\left(\mathbf{z}_t; \rho, \tau\right) - \epsilon \right)\frac{\partial{\mathbf{x}}}{\partial{\theta}}\right],
\end{equation}
where $\tau$ is timestamps in diffusion process, $\epsilon$ denotes noise, and $\omega$ is a weighted function. 
Intuitively, this loss perturbs $\mathbf{x}$ with a random amount of noise corresponding to the timestep $\tau$, and estimates an update direction that follows the score function of the diffusion model to move to a higher density region.

Besides text-to-3D, SDS is also widely used in image-to-3D tasks. 
Zero123~\citep{liu2023zero} is one prominent representative. 
It proposes a viewpoint-conditioned image-to-image translation diffusion model fine-tuned from Stable Diffusion~\citep{rombach2022high}, and exploits this 3D-aware image diffusion model to optimize a NeRF using SDS loss.
This image diffusion model takes one image, denoted by $\mathbf{I}_{in}$, and relative camera extrinsic between target view and input view, denoted by $\left(\mathbf{R}, \mathbf{T}\right)$, as the input, and outputs the target view image.
Compared with the original text-to-image diffusion model, text prompt in Equation~\ref{eq: sds} is not required in this model cause the authors translate the input image and the relative camera extrinsic to text embeddings. Then Equation~\ref{eq: sds} could be re-written as:
\begin{equation}\label{eq: zero123-sds}
   \nabla_{\theta}\mathcal{L}_{SDS}\left(\phi, \mathbf{x}\right) =  \mathbb{E}_{\tau, \epsilon} \left[\omega(t) \left( \hat{\epsilon}_{\theta}\left(\mathbf{z}_t; \mathbf{I}_{in}, \mathbf{R}, \mathbf{T}, \tau\right) - \epsilon \right)\frac{\partial{\mathbf{x}}}{\partial{\theta}}\right],
\end{equation}

\subsection{K-planes}
K-planes~\citep{fridovich2023k} is a simple and effective dynamic NeRF method which factorizes a dynamic 3D volume into six feature planes (i.e., hex-plane), denoted as $P=\{P_{o}\}$, where $o\in\{xy, yz, xz, xt, yt, zt\}$.  
The first three planes correspond to spatial dimensions, while the last three planes capture spatiotemporal variations. 
Each of the planes is structured as a $M \times M \times F$ tensor in the memory, where $M$ represents the size of the plane and $F$ is the feature size that encodes scene density and color information. Let $t$ denote the timestamp of a video clip, given a point $p = (x, y, z, t)$ in the 4D space, we normalize the coordinate to the range $[0, M)$ and subsequently project it onto the six planes using the equation $f(p)_o = P_o(\iota_o(p))$, where $\iota_o$ is the projection from a space point $p$ to a pixel on the $o$'th plane. The plane feature $f(p)_o$ is extracted via bilinear interpolation. 
The six plane features are combined using the Hadamard product (i.e., element-wise multiplication), to produce a final feature vector as follows:
\begin{equation}\label{eq: kplanes-multiply}
    f(p) = \prod_{o\in\{xy, yz, xz, xt, yt, zt\}} f(p)_o, 
\end{equation}
Then, the color and density of $p$ is calculated as $c(p) = c(f(p))$ and $d(p) = d(f(p))$, where $c$ and $d$ denotes mlps for color and density, respectively.

\section{Method}
\begin{figure}
    \centering
    \includegraphics[width=\textwidth]{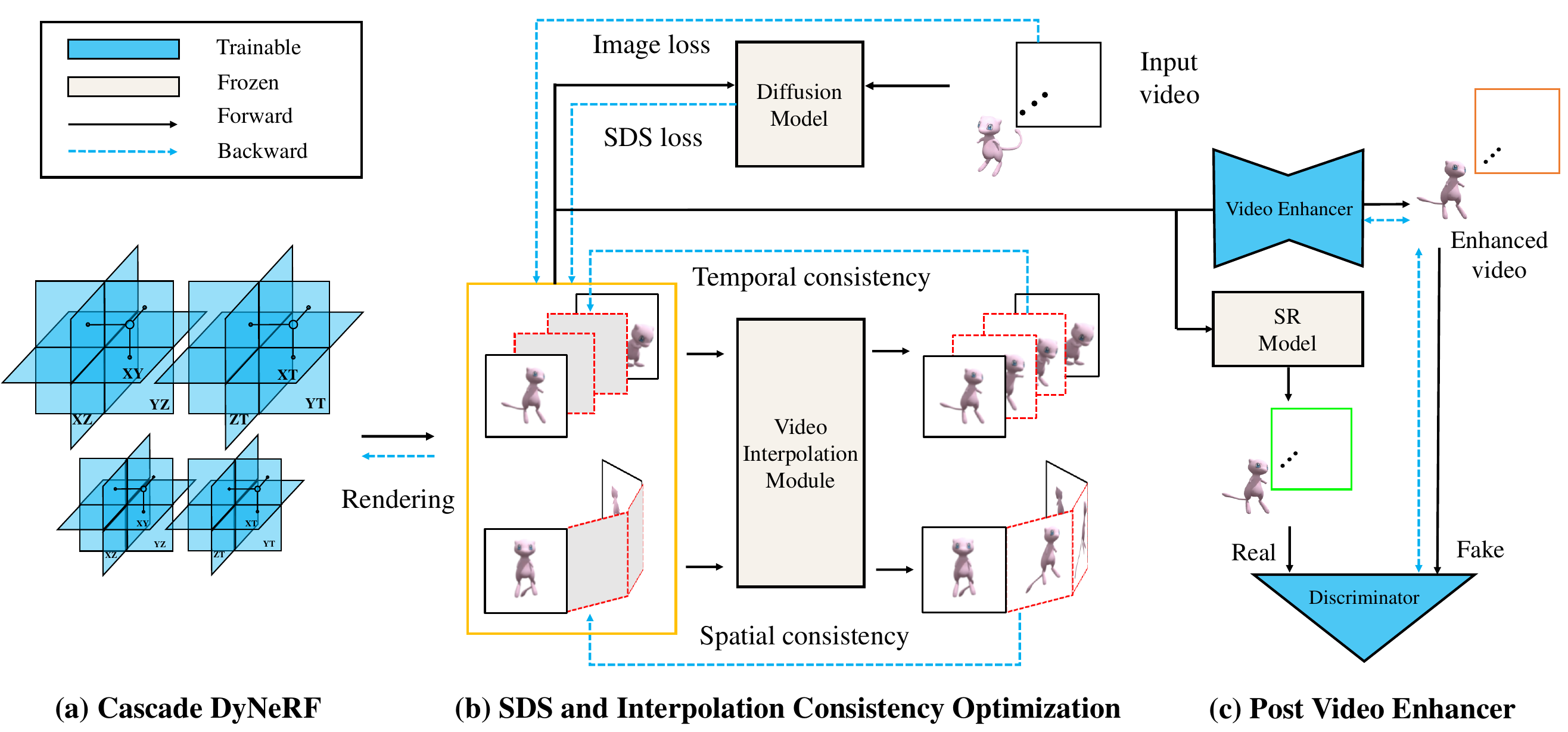}
    \caption{Schematic illustration of Consistent4D. 
    The framework consists of the optimization of a Cascade DyNeRF and the training of a post video enhancer. The Cascade DyNeRF, which adopts a residual learning strategy, is supervised by SDS loss from an image-to-image diffusion model. Particularly, a novel Interpolation-driven Consistency Loss is proposed to compensate for the spatiotemporal inconsistency brought by the image diffusion model. For post-processing, we train a lightweight cross-frame video enhancer using GAN to further improve the quality of the video rendered from DyNeRF.}
    \label{fig:pipeline}
\end{figure}
In this work, we target to generate a 360$^{\circ}$ dynamic object from a statically captured monocular video. 
To achieve this goal, we develop a framework consisting of a DyNeRF and a video enhancer, supervised by the pre-trained 2D diffusion model in Zero123 and a GAN, respectively.
As shown in Figure~\ref{fig:pipeline}, we first train a specially designed cascade DyNeRF using SDS loss and image reconstruction loss. To guarantee spatial and temporal consistency, we propose a novel Interpolation-driven Consistency Loss as the extra regularization for the DyNeRF.
For post-processing, we apply GAN to train a lightweight video enhancer, inspired by pix2pix~\citep{pix2pix2017}.To summarize, taking an uncalibrated monocular video as the input, we obtain a DyNeRF from which we can render 360$^{\circ}$ view of the dynamic object, and the rendered results can be further enhanced by the video enhancer. 

In the following sections, we will first introduce our design of the Cascade DyNeRF, and then illustrate the Interpolation-driven Consistency loss.  
At last, we will detail the video enhancer.

\subsection{Cascade DyNeRF}
Existing DyNeRF methods mainly assume the supervision signals are temporally coherent, however, this assumption does not hold in our task since our main supervision signal is from an image diffusion model. 
In order to minimize the impact of temporal discontinuity in the supervision signals, we are prone to 4D representations which naturally guarantee a certain level of temporal continuity. 
Therefore, we build our DyNeRF based on K-planes~\citep{fridovich2023k} which explots temporal interpolation, an operator naturally inclined to temporal smoothing.
Empirically, reducing the time resolution of spatiotemporal planes helps enhance temporal consistency, however, this results in over-smoothed renderings where finer details are lost. In contrast, increasing the time resolution leads to crisp renderings, but the continuity of images within the same time series diminishes. To achieve both temporal continuity and high image quality, we adjust the multi-scale technique in K-planes and introduce Cascade DyNeRF.

Let us denote the scale index by $s$. In K-planes, multi-scale features are exploited by concatenation along feature dimension, then the color and density could be calculated as:
\begin{equation}\label{eq: kplanes-concat}
        c(p) = c(\mathrm{concat}(\{f(p)^s)\}_{s=1}^S), \;
        d(p) = d(\mathrm{concat}(\{f(p)^s)\}_{s=1}^S),
\end{equation}
where $S$ is the number of scales.
In our setting, the experiment results show simple concatenation is hard to balance between image quality and temporal consistency. So we propose to leverage the cascade architecture, which is commonly used in object detection algorithms~\citep{cai2018cascade, carion2020end}, to first output a coarse yet temporally coherent dynamic object from low-resolution planes and then let the high-resolution planes learn the residual between the coarse and fine results. That is, the color and density of the object after scale $s$ is: 
\begin{equation}\label{eq: kplanes-cascade}
        c(p)^s = \sum_{k=1}^s c(f(p)^k), \;
        d(p)^s = \sum_{k=1}^s d(f(p)^k),
\end{equation}
where k indicates the scale index.
Note that losses are applied to the rendering results of each scale to guarantee that planes with higher resolution learn the residual between results from previous scales and the target object.
In this way, we can improve temporal consistency without sacrificing much object quality. 
But Cascade DyNeRF alone is not enough for spatiotemporal consistency, so we resort to extra regularization, please see in the next section.

\subsection{Interpolation-driven Consistency Loss}
Video generation methods usually train an inter-frame interpolation module to enhance the temporal consistency between keyframes~\citep{ho2022imagen, zhou2022magicvideo, blattmann2023align}. 
Inspired by this, we exploit a pre-trained light-weighted video interpolation model and propose Interpolation-driven Consistency Loss to enhance the spatiotemporal consistency of the 4D generation.

The interpolation model adopted in this work is RIFE~\citep{huang2022rife}, which takes a pair of consecutive images as well as the interpolation ratio $\gamma ~ (0 < \gamma < 1)$ as the input, and outputs the interpolated image. 
In our case, we first render a batch of images that are either spatially continuous or temporally continuous, denoted by $\{\mathbf{x}\}_{j=1}^{J}$, where $J$ is the number of images in a batch. Let us denote the video interpolation model as $\psi$, the interpolated image as $\hat{\mathbf{x}}$, then we calculate the Interpolation-driven Consistency Loss as:
\begin{equation}\label{eq: loss-icl}
    \begin{aligned}
        \hat{\mathbf{x}}_j  &= \psi(\mathbf{x}_1, \mathbf{x}_J, \gamma_j ), \\
        \mathcal{L}_{ICL} &=  \sum_{j=2}^{J-1} \lVert \mathbf{x}_j - \hat{\mathbf{x}}_j \rVert_2,
    \end{aligned}
\end{equation}
where $\gamma_{j} =\frac{j-1}{J-1}$, and $2 \le j \le J-1$. 

This simple yet effective loss enhances the continuity between frames thus improving the spatiotemporal consistency in dynamic object generation by a large margin. Moreover, we find the spatial version of this loss alleviates the multi-face problem in 3D generation tasks as well. Please refer to the experiment sections to see quantitative and qualitative results. The Interpolation-driven Consistency Loss and some other regularization losses are added with SDS loss in Equation ~\ref{eq: zero123-sds}, details of which can be found in the experiment section.  

\subsection{Cross-frame Video Enhancer}
Sometimes image sequence rendered from the optimized DyNeRF suffers from artifacts, such as blurry edges, small floaters, and insufficient smoothness, especially when the object motion is abrupt or complex. To further improve the quality of rendered videos, we design a lightweight video enhancer and optimize it via GAN, following pix2pix~\citep{pix2pix2017}. 
The real images are obtained with image-to-image technique~\citep{meng2021sdedit} using a super-resolution diffusion model, and the fake images are the rendered ones.

To better exploit video information, We add cross-frame attention to the UNet architecture in pix2pix, i.e., each frame will query information from two adjacent frames. We believe this could enable better consistency and image quality. Denote the feature map before and after cross-frame-attention as $\mathbf{F}$ and $\mathbf{F}_j'$, we have:
\begin{equation}
    \begin{aligned}
        F_j'&=\mathrm{Attention}(\mathcal{Q}_j, \mathcal{K}_j, \mathcal{V}_j), \\
        \mathcal{Q}_j=\mathrm{flatten}(F_j),\;& \mathcal{K}_j=\mathcal{V}_j=\mathrm{flatten}(\mathrm{concat}(F_{j-1}, F_{j+1}),
    \end{aligned}
\end{equation}
where $\mathcal{Q}$, $\mathcal{K}$ and $\mathcal{V}$ denotes query, key, and value in attention mechanism, and $\mathrm{concat}$ denotes the concatenation along the width dimension.

Loss for the generator and discriminator are the same as pix2pix.

\section{Experiment}
We have conducted extensive experiments to evaluate the proposed Consistent4D generator using both synthetic data and in-the-wild data. The experimental setup, comparison with dynamic NeRF baselines, and ablations are provided in the following sections.

\subsection{Implementation Details}
\textbf{Data Preparation} For each input video, we initially segment the foreground object utilizing SAM~\citep{kirillov2023segment} and subsequently sample 32 frames uniformly. The majority of the input videos span approximately 2 seconds, with some variations extending to around 1 second or exceeding 5 seconds. For the ablation study of video sampling, please refer to the appendix~\ref{sec:num_frames}.

\textbf{Training} During SDS and interpolation consistency optimization, we utilize zero123-xl trained by ~\cite{deitke2023objaverse} as the diffusion model for SDS loss. For Cascade DyNeRF, we set $s=2$, i.e., we have coarse-level and fine-level DyNeRFs.The spatial and temporal resolution of Cascade DyNeRF are configured to 50 and 8 for coarse-level, and 100 and 16 for fine-level, respectively. We first train DyNeRF with batch size 4 and resolution 64 for 5000 iterations. Then we decrease the batch size to 1 and increase the resolution to 256 for the next 5000 iteration training.
ICL is employed in the initial 5000 iterations with a probability of 25\%, and we sample consecutive temporal frames at intervals of one frame and sample consecutive spatial frames at angular intervals of 5$^{\circ}$-15$^{\circ}$ in azimuth.
SDS loss weight is set as 0.01 and reconstruction loss weight is set as 500.
In addition to SDS and ICL, we also apply foreground mask loss, normal orientation loss, and 3D smoothness loss. 
The learning rate is set as 0.1 and the optimizer is Adam.
In the post-video enhancing stage, we train the video enhancer with a modified Unet architecture. The learning rate is set as 0.002, the batch size is 16, and the training epoch is 200.
The main optimization stage and the video enhancing stage cost about 2.5 hours and 15 minutes on a single V100 GPU.
For details, please refer to the appendix~\ref{sec:extra_imple}.

\begin{table}[t]
\centering
\resizebox{\textwidth}{!}{%
\begin{tabular}{c cc cc cc cc cc cc cc cc}
\specialrule{.15em}{.1em}{.1em}
& \multicolumn{2}{c}{\textit{Pistol}} & \multicolumn{2}{c}{\textit{Guppie}} 
& \multicolumn{2}{c}{\textit{Crocodile}} & \multicolumn{2}{c}{\textit{Monster}} 
& \multicolumn{2}{c}{\textit{Skull}} & \multicolumn{2}{c}{\textit{Trump}} & \multicolumn{2}{c}{\textit{Aurorus}} 
& \multicolumn{2}{c}{Average} \\ 
& LPIPS & CLIP & LPIPS & CLIP & LPIPS & CLIP & LPIPS & CLIP & LPIPS & CLIP & LPIPS & CLIP & LPIPS & CLIP & LPIPS & CLIP \\ 
\specialrule{.15em}{.1em}{.1em}
D-NeRF & 0.52 & 0.66 & 0.32 & 0.76 & 0.54 & 0.61 & 0.52 & 0.79 & 0.53 & 0.72 & 0.55 & 0.60 & 0.56 & 0.66 & 0.51 & 0.68\\
K-planes & 0.40 & 0.74 & 0.29 & 0.75 & 0.19 & 0.75 & 0.47 & 0.73 & 0.41 & 0.72 & 051 & 0.66 & 0.37 & 0.67 & 0.38 & 0.72 \\
ours & \textbf{0.10} & \textbf{0.90} & \textbf{0.12} & \textbf{0.90} & \textbf{0.12} & \textbf{0.82} & \textbf{0.18} & \textbf{0.90} & \textbf{0.17} & \textbf{0.88} & \textbf{0.23} & \textbf{0.85} & \textbf{0.17} & \textbf{0.85} & \textbf{0.16} & \textbf{0.87} \\
\specialrule{.15em}{.1em}{.1em}
\end{tabular}%
}
\caption{Video-to-4D quantitative comparison.}
\label{tab:video-to-4d-quan}
\end{table}
\begin{figure}[!htb]
    \centering
    \includegraphics[width=0.85\textwidth]{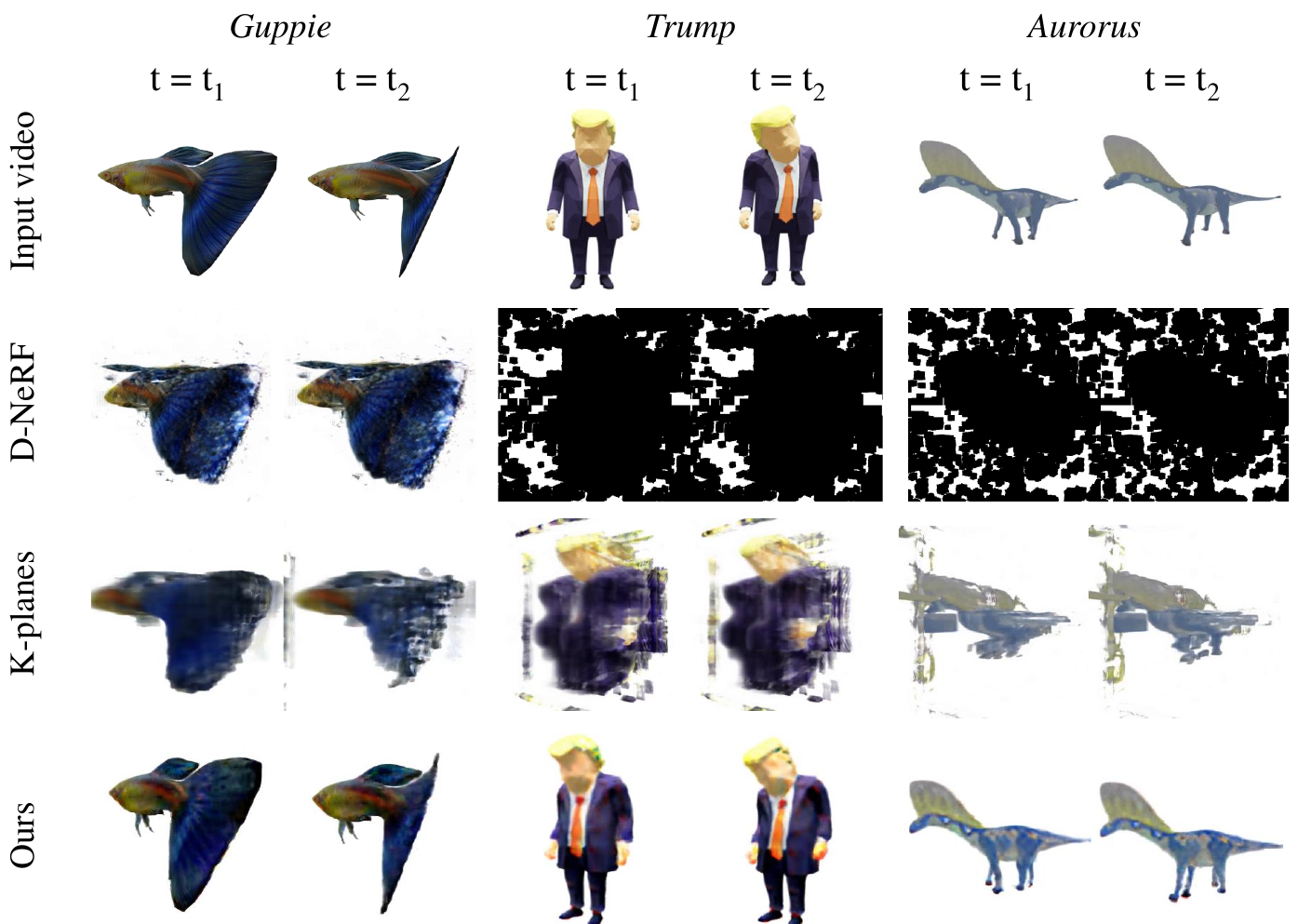}
    \caption{Comparison with dynamic NeRF methods. We render each dynamic object from a novel view at two timestamps.}
    \label{fig:comparison}
\end{figure}
\subsection{Comparisons with Other Methods}
To date, few methods have been developed for 4D generation utilizing video obtained from a static camera, so we only manage to compare our method with D-NeRF~\citep{pumarola2021d} and K-planes~\citep{fridovich2023k}.

\textbf{Quantitative Results}
To quantitatively evaluate the proposed video-4D generation method, we select and download seven animated models, namely \textit{Pistol}, \textit{Guppie}, \textit{Crocodie}, \textit{Monster}, \textit{Skull}, \textit{Trump}, \textit{Aurorus}, from Sketchfab~\citep{sketchfab2023} and render the multi-view videos by ourselves, as shown in Figure~\ref{fig:comparison} and appendix~\ref{sec:rest_data}. We render one input view for scene generation and 4 testing views for our evaluation. The per-frame LPIPS~\citep{zhang2018unreasonable} score and the CLIP~\citep{radford2021learning} similarity are computed between testing and rendered videos. We report the scores averaged over the four testing views in Table.~\ref{tab:video-to-4d-quan}.
Note that the commonly used PSNR and SSIM scores were not applied in our scenario as the pixel- and patch-wise similarities are too sensitive to the reconstruction difference, which does not align with the generation quality usually perceived by humans.
As shown in Table ~\ref{tab:video-to-4d-quan}, our dynamic 3D generation produces the best quantitative results over the other two methods on both the LPIPS and CLIP scores, which well aligns with the qualitative comparisons shown in Figure ~\ref{fig:comparison}.

\textbf{Qualitative Results}
The outcomes of our method and those of dyNeRFs are illustrated in Figure~\ref{fig:comparison}. It is observable that both D-NeRF and HyperNeRF methods struggle to achieve satisfactory results in novel views, owing to the absence of multi-view information in the training data. In contrast, leveraging the strengths of the generation model, our method proficiently generates a 360$^{\circ}$ representation of the dynamic object. For additional results, please refer to the appendix~\ref{sec:more_results}.

\subsection{Ablations} 
We perform ablation studies for every component within our framework. For clarity, the video enhancer is excluded when conducting ablations for SDS and interpolation consistency optimization.
\begin{figure}[!htb]
    \centering
    \includegraphics[width=0.9\textwidth]{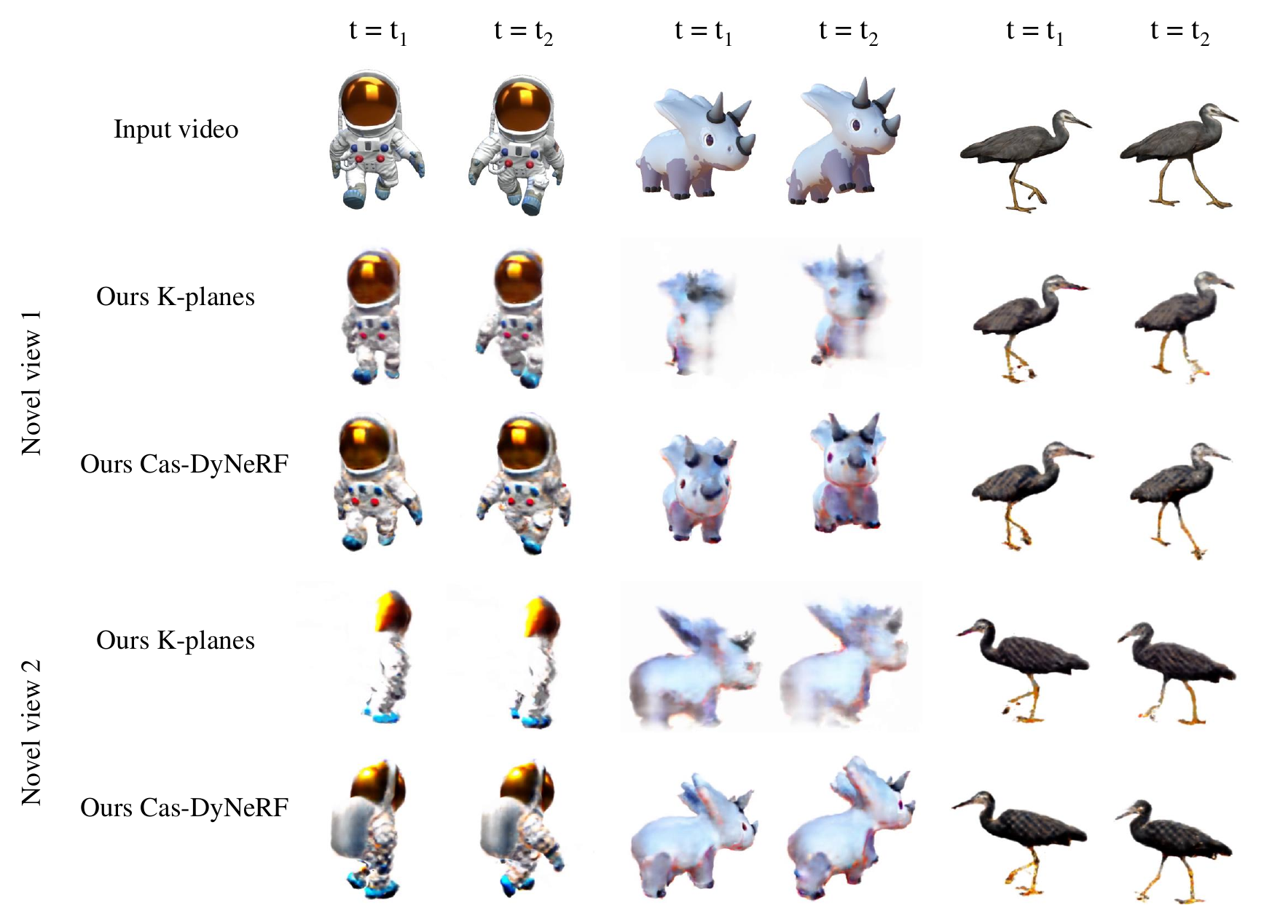}
    \caption{Ablation of Cascade DyNeRF.}
    \label{fig:ablation-cascade-dynerf}
\end{figure}
\begin{figure}[!htb]
  \centering
  \begin{subfigure}{\textwidth}
    \centering
    \includegraphics[width=\textwidth]{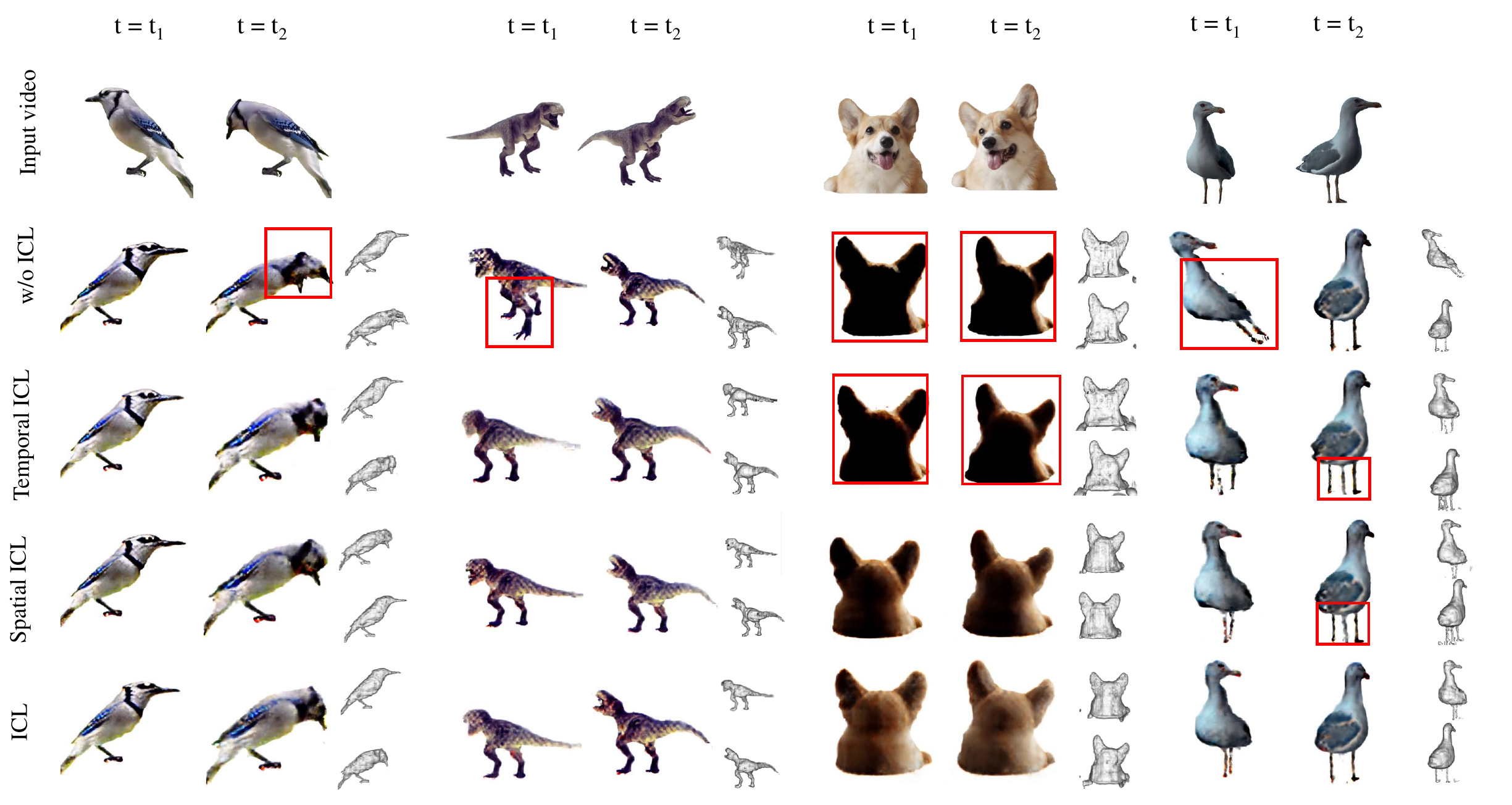}
    \caption{Video-to-4D. For each dynamic object, we render it from a novel view for two timestamps with textureless rendering for each timestamp. For clarity, we describe the input videos as follows (from left to right): \textit{blue jay pecking}, \textit{T-rex roaring}, \textit{corgi smiling}, \textit{seagull turning around}.}
    \label{fig:ablation-video-to-4d}
  \end{subfigure}
  \vfill
  \begin{subfigure}{\textwidth}
    \centering
    \includegraphics[width=\textwidth]{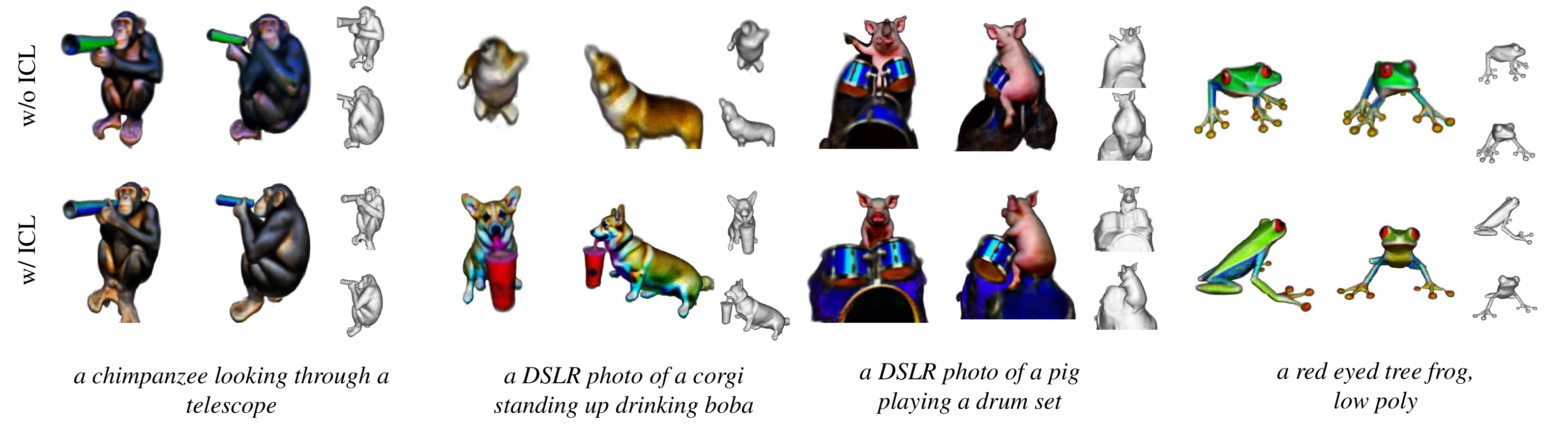}
    \caption{Text-to-3D. For each 3D model, we render it from two views with a textureless rendering for each view and remove the background to focus on the actual 3D shape.}
    \label{fig:ablation-text-to-3d}
  \end{subfigure}
  \caption{Ablation of Interpolation-driven Consistency Loss.}
  \label{fig:ablation-vif}
\end{figure}
\begin{table}[!htb]
  \centering
  \begin{minipage}[b]{0.45\textwidth}
    \centering
    \begin{subtable}{\textwidth}
      \centering
      \begin{tabular}{ccc}
        \specialrule{.15em}{.1em}{.1em}
        & w/o ICL & w/ ICL \\
        \hline
        preference rate(\%) & 24.5 & \textbf{75.5} \\
        \specialrule{.15em}{.1em}{.1em}
      \end{tabular}
      \caption{Video-to-4D.}
      \label{tab:user-study-video-to-4d}
    \end{subtable}
    \vfill
    \begin{subtable}{\textwidth}
      \centering
      \begin{tabular}{ccc}
        \specialrule{.15em}{.1em}{.1em}
        & w/o ICL & w/ ICL \\
        \hline
        success rate(\%) & 19.3 & \textbf{28.6} \\
        \specialrule{.2em}{.1em}{.1em}
      \end{tabular}
      \caption{Text-to-3D.}
      \label{tab:user-study-text-to-3d}
    \end{subtable}
  \end{minipage}
  \hfill
  \begin{minipage}[b]{0.525\textwidth}
    \centering
    \includegraphics[width=1.\textwidth, trim={0.9cm 1.2cm 0.9cm 1cm}, clip]{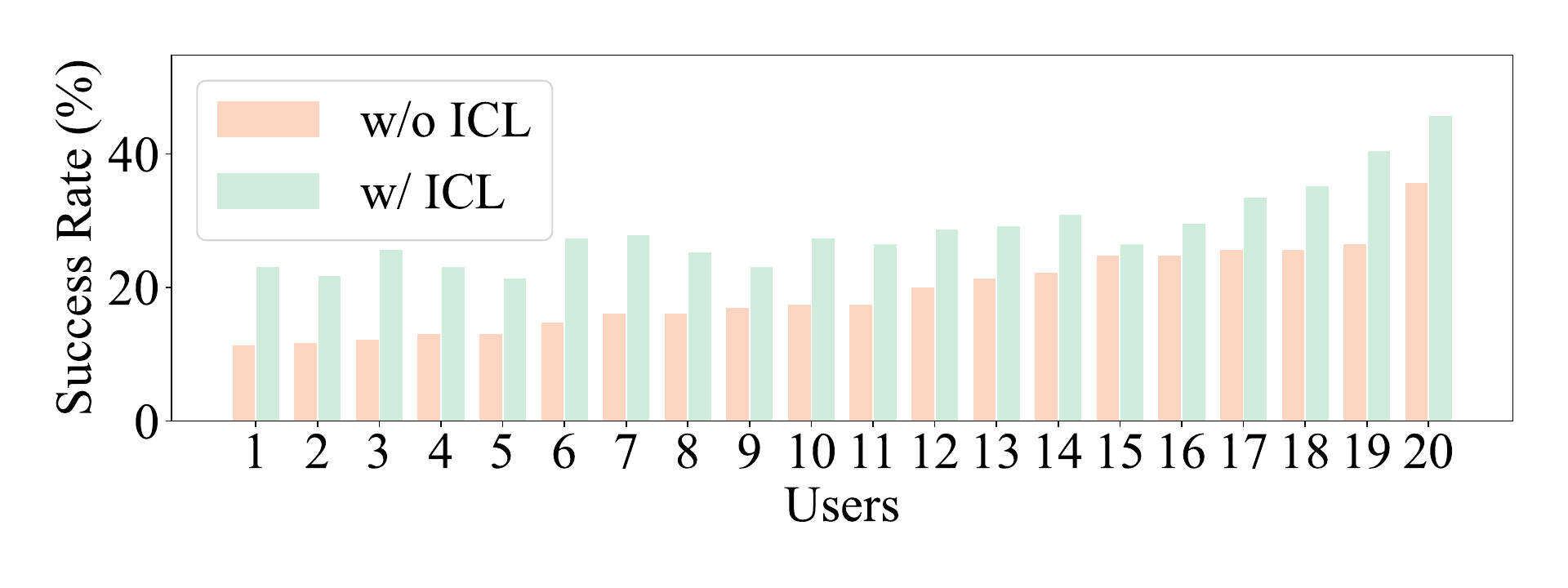}
    \captionsetup{skip=-7pt} 
    \refstepcounter{subtable}
    \caption*{(c) Details of Text-to-3D user study. Data is sorted.}
    \label{tab:user-study-text-to-3d-details}
  \end{minipage}
  \caption{User study of Interpolation-driven Consistency Loss.}
  \label{tab:user-study}
\end{table}

\textbf{Cascade DyNeRF}
In Figure~\ref{fig:ablation-cascade-dynerf}, we conduct an ablation study for Cascade DyNeRF. Specifically, we substitute Cascade DyNeRF with the original K-planes architecture, maintaining all other settings unchanged.
In the absence of the cascade architecture, the training proves to be unstable, occasionally yielding incomplete or blurry objects, as demonstrated by the first and second objects in Figure~\ref{fig:ablation-cascade-dynerf}. In some cases, while the model manages to generate a complete object, the moving parts of the object lack clarity, exemplified by the leg and beak of the bird.
Conversely, the proposed Cascade DyNeRF exhibits stable training, leading to relatively satisfactory generation results.

\textbf{Interpolation-driven Consistency Loss}
The introduction of Interpolation-driven Consistency Loss (ICL) stands as a significant contribution of our work. Therefore, we conduct extensive experiments to investigate both its advantages and potential limitations. 
Figure~\ref{fig:ablation-video-to-4d} illustrates the ablation of both spatial and temporal Interpolation-driven Consistency Loss (ICL) in the video-to-4D task. 
Without ICL, the objects generated exhibit spatial and temporal inconsistency, as evidenced by the multi-face/foot issue in the blue jay and T-rex, and the unnatural pose of the seagull. Additionally, color discrepancies, such as the black backside of the corgi, are also noticeable.
Employing either spatial or temporal ICL mitigates the multi-face issue, and notably, the use of spatial ICL also alleviates the color defect problem. Utilizing both spatial and temporal ICL concurrently yields superior results.
We further perform a user study, depicted in Figure~\ref{tab:user-study-video-to-4d}, which includes results w/ and w/o ICL for 20 objects. For efficiency in evaluation, cases in which both methods fail are filtered out in this study. 20 users participate in this evaluation, and the results unveiled a preference for results w/ ICL in 75\% of the cases.

We further explore whether ICL could alleviate multi-face problems for text-to-3D tasks. We compared the success rate of DreamFusion implemented w/ and w/o the proposed ICL loss. 
For the sake of fairness and rigor, we collect all prompts related to animals from the official DreamFusion project page, totaling 230. 
20 users are asked to participate in this
\texttt{non-cherry-pick} user study, where we establish three criteria for a successful generation: \texttt{alignment with the text prompt}, \texttt{absence of multi-face issues}, and \texttt{clarity in geometry and texture}. We visualize the statistics in Table~\ref{tab:user-study-text-to-3d} and Table~\ref{tab:user-study-text-to-3d-details}. The results show that although users have different understandings of successful generation, results w/ ICL always outperform results w/o it.
For a comprehensive understanding, qualitative comparisons are presented in Figure~\ref{fig:ablation-text-to-3d}, which indicates the proposed technique effectively alleviates the multi-face Janus problem and thus promotes the success rate. Implementation details about text-to-3D can be found in the appendix~\ref{sec:extra_imple}, and we also analyze the failure cases and limitations of ICL in~\ref{sec:fail_case}.

\textbf{Cross-frame Video Enhancer}
In Figure~\ref{fig:ablation-video-enhancer} (see in appendix), we show the proposed cross-frame video enhancer could improve uneven color distribution and smooth out the rough edges, as shown in almost all figures, and remove some floaters, as indicated by the cat in the red and green box.

\section{Conclusion}
We introduce a novel video-to-4D framework, named Consistent4D, aimed at generating 360$^{\circ}$ 4D objects from uncalibrated monocular videos captured by a stationary camera. Specifically, we first optimize a Cascade DyNeRF which is specially designed to facilitate stable training under the discrete supervisory signals from an image-to-image diffusion model. More crucially, we introduce an Interpolation-driven Consistency Loss to enhance spatial and temporal consistency, the main challenge in this task. For comprehensiveness, we train a lightweight video enhancer to rectify scattered color discrepancies and eliminate minor floating artifacts, as a post-processing step. Extensive experiments conducted on both synthetic and in-the-wild data demonstrate the effectiveness of our method.

\bibliography{iclr2024_conference}
\bibliographystyle{iclr2024_conference}

\newpage
\appendix
\section{Appendix}
\subsection{Additional Visualization Results}\label{sec:more_results}
\begin{figure}[tbp]
    \centering
    \includegraphics[width=\textwidth]{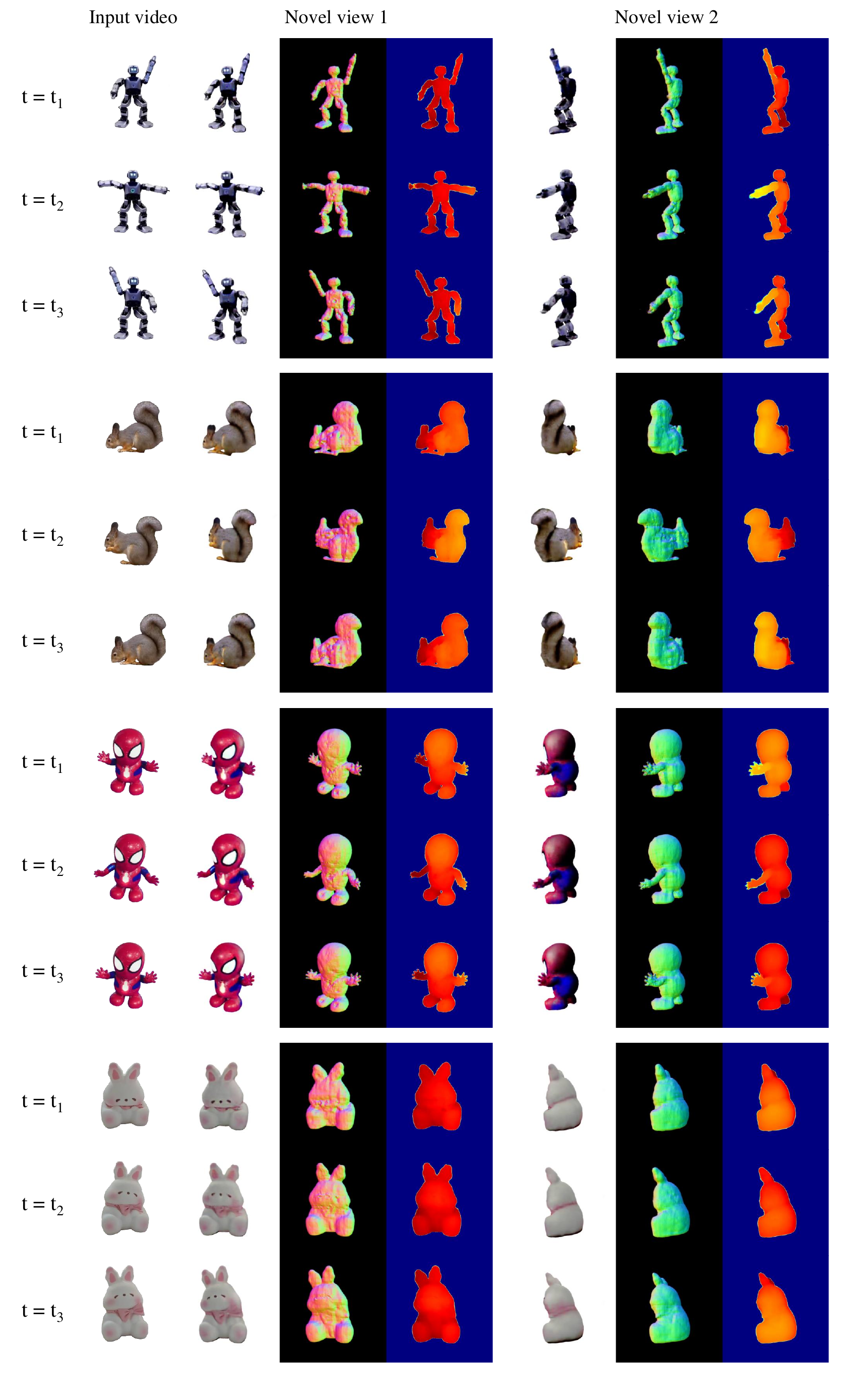}
    \caption{Visualization results of our method. All four input videos are in-the-wild videos. The novel views presented are 22.5$^{\circ}$ and 112.5$^{\circ}$ away from the input view, respectively. The results of our methods include RGB, normal map and depth map (from left to right).}
    \label{fig:more_results}
\end{figure}
In Figure~\ref{fig:more_results}, we present the result of our method on four in-the-wild videos. For clarity, we describe the input videos as follows: \textit{robot dancing}, \textit{squirrel feeding}, \textit{toy-spiderman dancing}, \textit{toy-rabbit deforming}. Due to limited space, the reviewers are strongly recommended to watch the video in the attached files to see various visualization results.
\subsection{Data Used in Video-to-4D Quantitative Evaluation}\label{sec:rest_data}
\begin{figure}
    \centering
    \includegraphics[width=\textwidth]{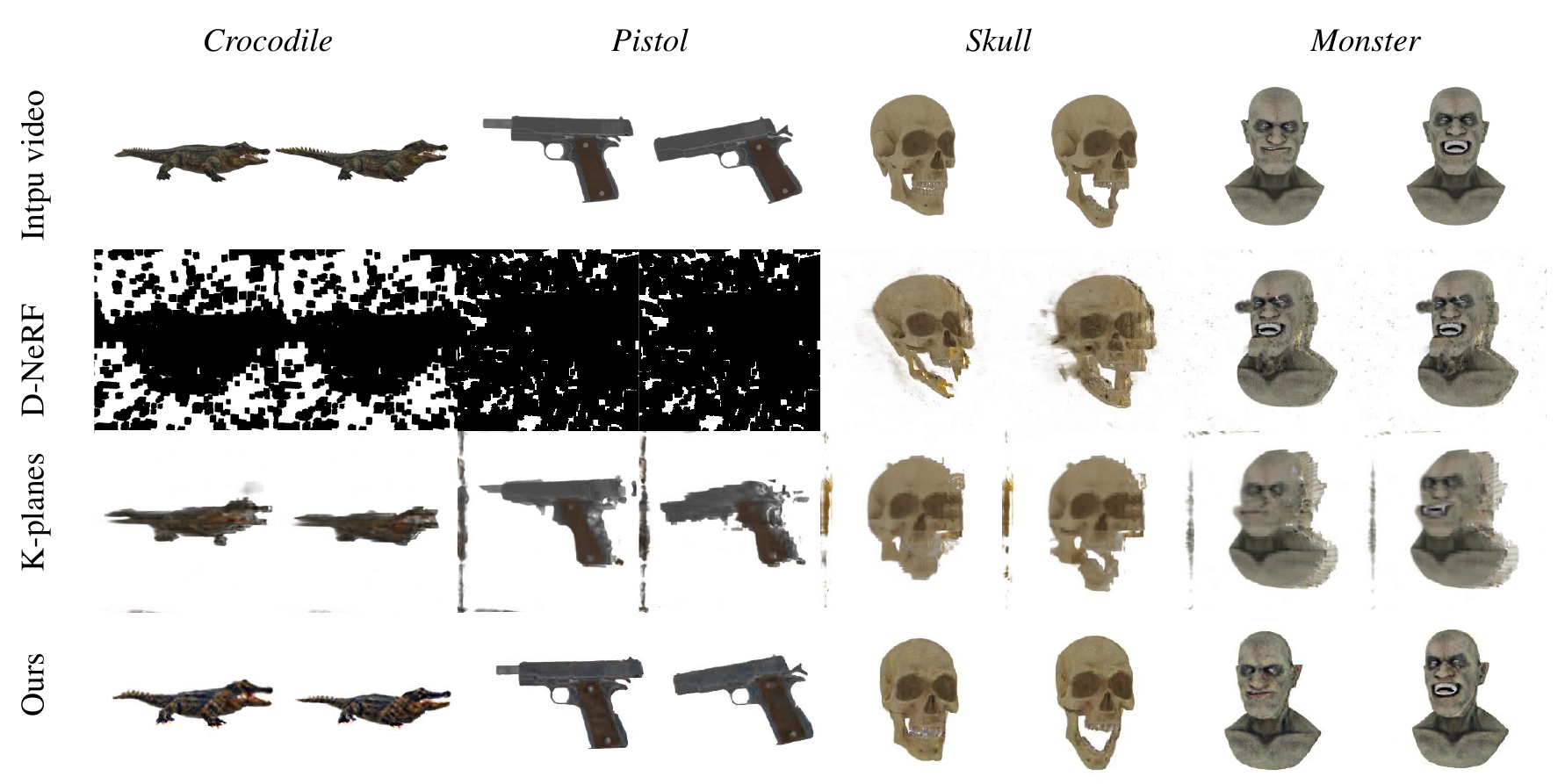}
    \caption{Data and comparison results for video-to-4D quantitative evaluation.}
    \label{fig:rest_data}
\end{figure}
Sin three dynamic objects are shown in Figure~\ref{fig:rest_data}, we only visualize the rest four here, as shown in Figure~\ref{fig:rest_data}. The observation is similar to the results in the main paper.

We demonstrate the effectiveness of video enhancer in Figure~\ref{fig:ablation-video-enhancer}. The analysis can be found in the main paper.

\subsection{The Number of Frames}\label{sec:num_frames}
\begin{figure}[tbp]
    \centering
    \includegraphics[width=\textwidth]{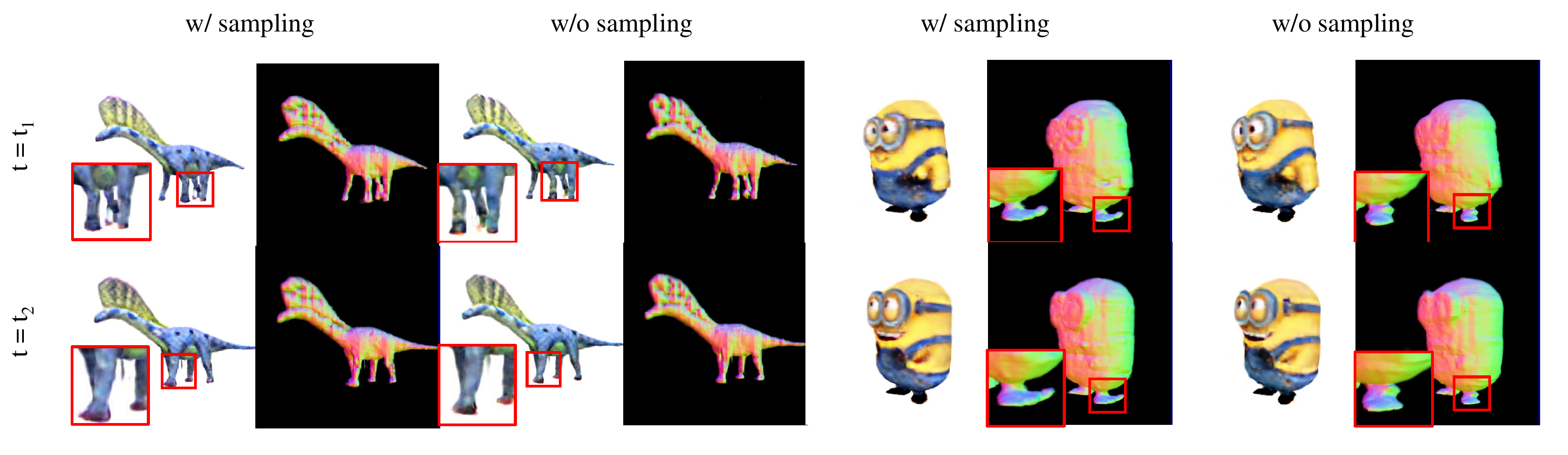}
    \caption{Ablation of video frame sampling. Videos w/ sampling contain 32 frames. Videos w/o sampling contain 72 and 39 frames for \textit{Aurorus} and \textit{minions}, respectively. The results of our methods include RGB and normal map (from left to right).}
    \label{fig:video_frames}
\end{figure}
For simplicity, we sample each input video to 32 frames in all experiments. However, we find input videos without sampling sometimes give slightly better results, as shown in Figure~\ref{fig:video_frames}.

\subsection{Implementation Details}\label{sec:extra_imple}
\textbf{Loss Function in Video-to-4D}
Besides SDS loss $\mathcal{L}_{SDS}$, Interpolation-driven consistency loss $\mathcal{L}_{ICL}$, we also apply reconstruction loss $\mathcal{L}_{rec}$ and mask loss $\mathcal{L}_{m}$ for the input view. 
3D normal smooth loss $\mathcal{L}_{n}$ and orientation loss $\mathcal{L}_{ori}$~\citep{verbin2022ref} are also applied. Therefore, the final loss is calculated as $\mathcal{L} = \lambda_1\mathcal{L}_{SDS} + \lambda_2\mathcal{L}_{ICL} + \lambda_3\mathcal{L}_{rec} + \lambda_4\mathcal{L}_{m} + \lambda_5\mathcal{L}_{n} + \lambda_6\mathcal{L}_{ori}$, where $\lambda_1=0.1$, $\lambda_2=2500$, $\lambda_3=500$, $\lambda_4=50$, $\lambda_5=2.0$, and $\lambda_6$ is initially 1 and increased to 20 linearly until 5000 iterations. Note that the reconstruction loss and SDS loss are applied alternatively.  

\begin{figure}[!htb]
    \centering
    \includegraphics[width=\textwidth]{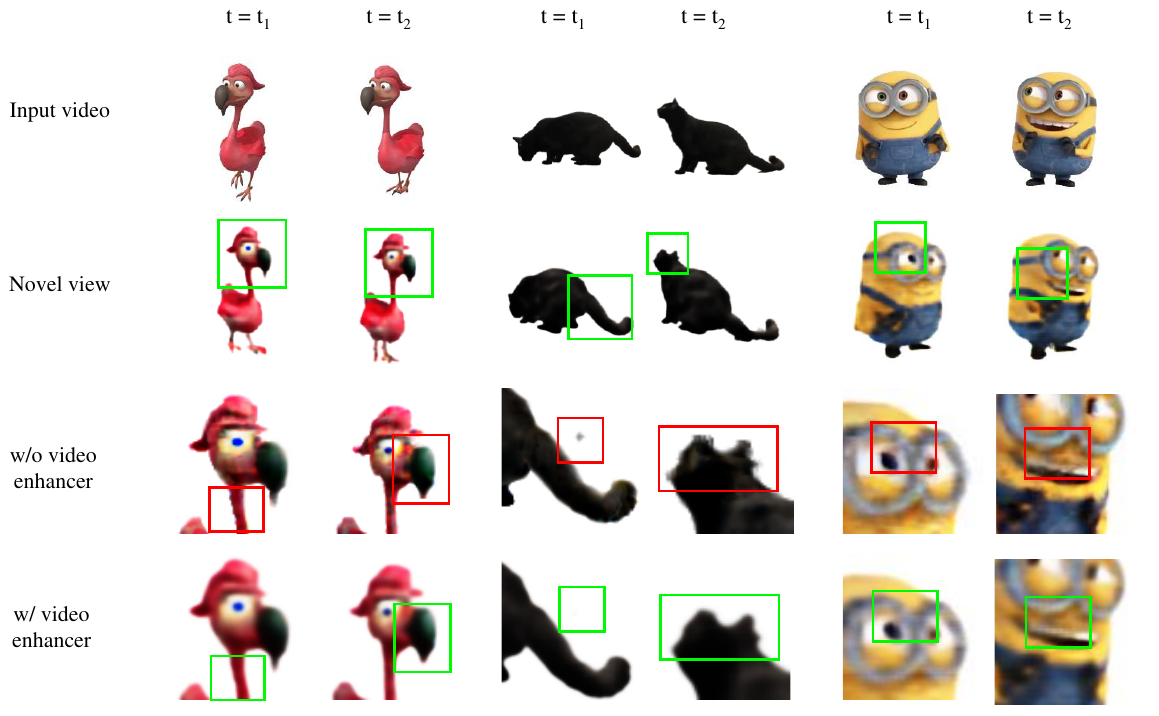}
    \caption{Ablation of Video enhancer. Please zoom in to view the details. }
    \label{fig:ablation-video-enhancer}
\end{figure}

\textbf{Video Enhancer}
For video enhancer architecture, we follow pix2pix~\citep{pix2pix2017} except for that we modify the unet256 architecture to a light-weighted version, with only three up/down layers and one cross-frame attention layer. The feature dimensions for the unet layers are set as 64, 128, and 256. Besides, we inject a cross-attention layer in the inner layer of the unet to enable the current frame to query information from adjacent frames.
For real images, we use DeepFloyd-IF stage II~\citep{deepfloyd2023}, which is a diffusion model for super-resolution tasks. The input image, i.e., the rendered image, is resized to $64\times 64$ and the output resolution is $256\times 256$. The prompt needed by the diffusion model is manually set, i.e., we use the "a $*$" as the prompt, in which $*$ is the category of the dynamic object. 
For example, the prompts for dynamic objects in Figure~\ref{fig:ablation-video-enhancer} are \textit{a bird}, \textit{a cat}, \textit{a minions}. 
The prompt cloud also be obtained from image or video caption models, or large language models.

\textbf{Text-to-3D Details}
We choose Threestudio built by ~\citep{threestudio2023} as the codebase since it is the best public implementation we could find. DeepFloy-IF~\citep{deepfloyd2023} is employed as the diffusion model, and all default tricks in Threestudio are utilized. The hyper-parameters for results w/ and w/o ICL, such as batch size and learning rate, are kept consistent between the implementations w/ and w/o ICL, except for those related to ICL.
We train the model for 5000 iterations, the first 1000 iterations with batch size 8 and resolution 64, and the rest 4000 with batch size 2 and resolution 256. The learning rate is 0.01 and the optimizer is Adam, the same as the default setting in Threestudio. 
The ICL loss is applied in the first 1000 iterations with probability 30\% and weight 2000.

\subsection{Failure Cases}\label{sec:fail_case}
\begin{figure}
    \centering
    \begin{minipage}{0.5\textwidth}
        \centering
        \includegraphics[width=\linewidth]{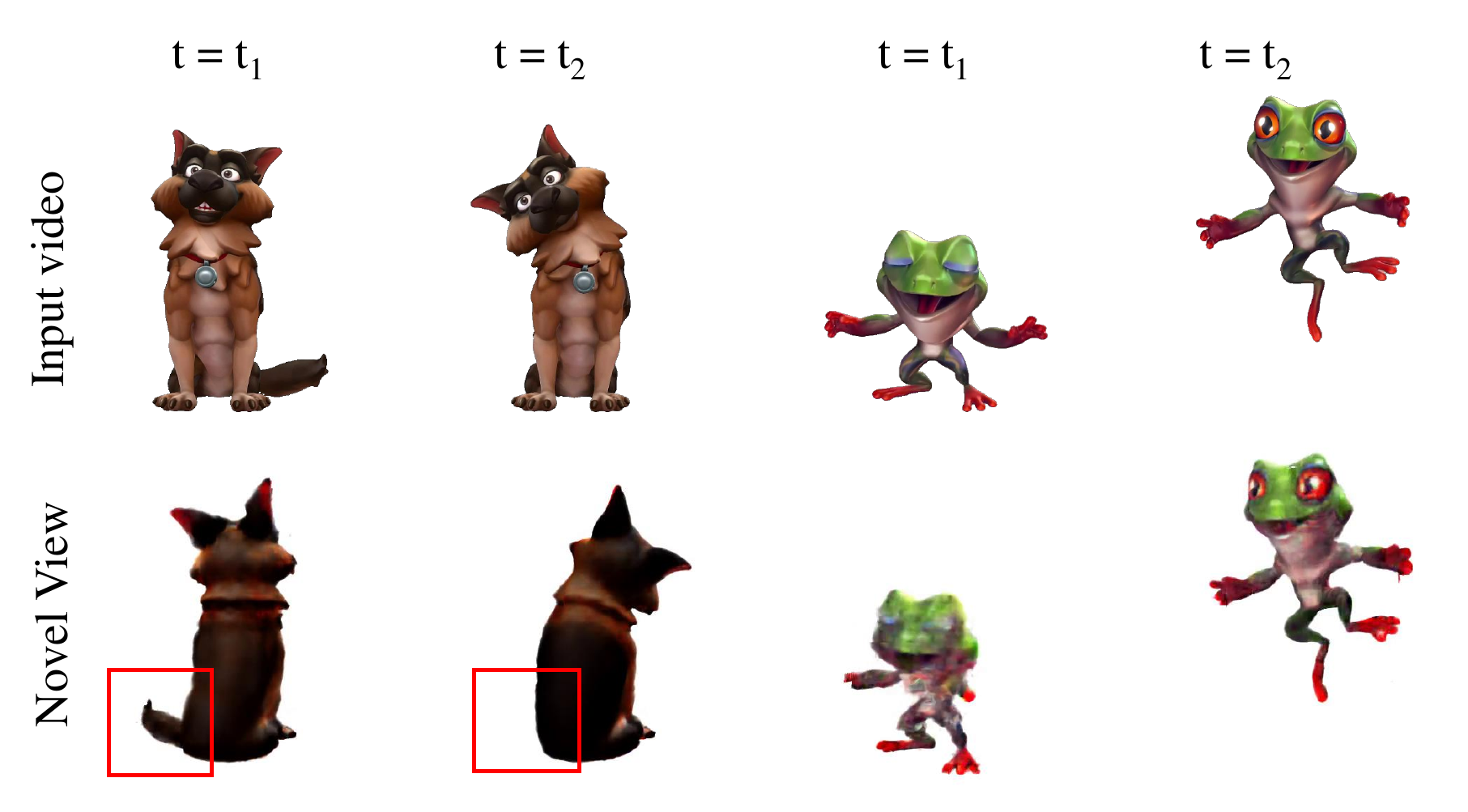}
        \vspace{0.01cm}
        \caption{Video-to-4D failure cases.}
        \label{fig:fail_4d}
    \end{minipage}
    \hfill
    \begin{minipage}{0.475\textwidth}
        \centering
        \includegraphics[width=\linewidth]{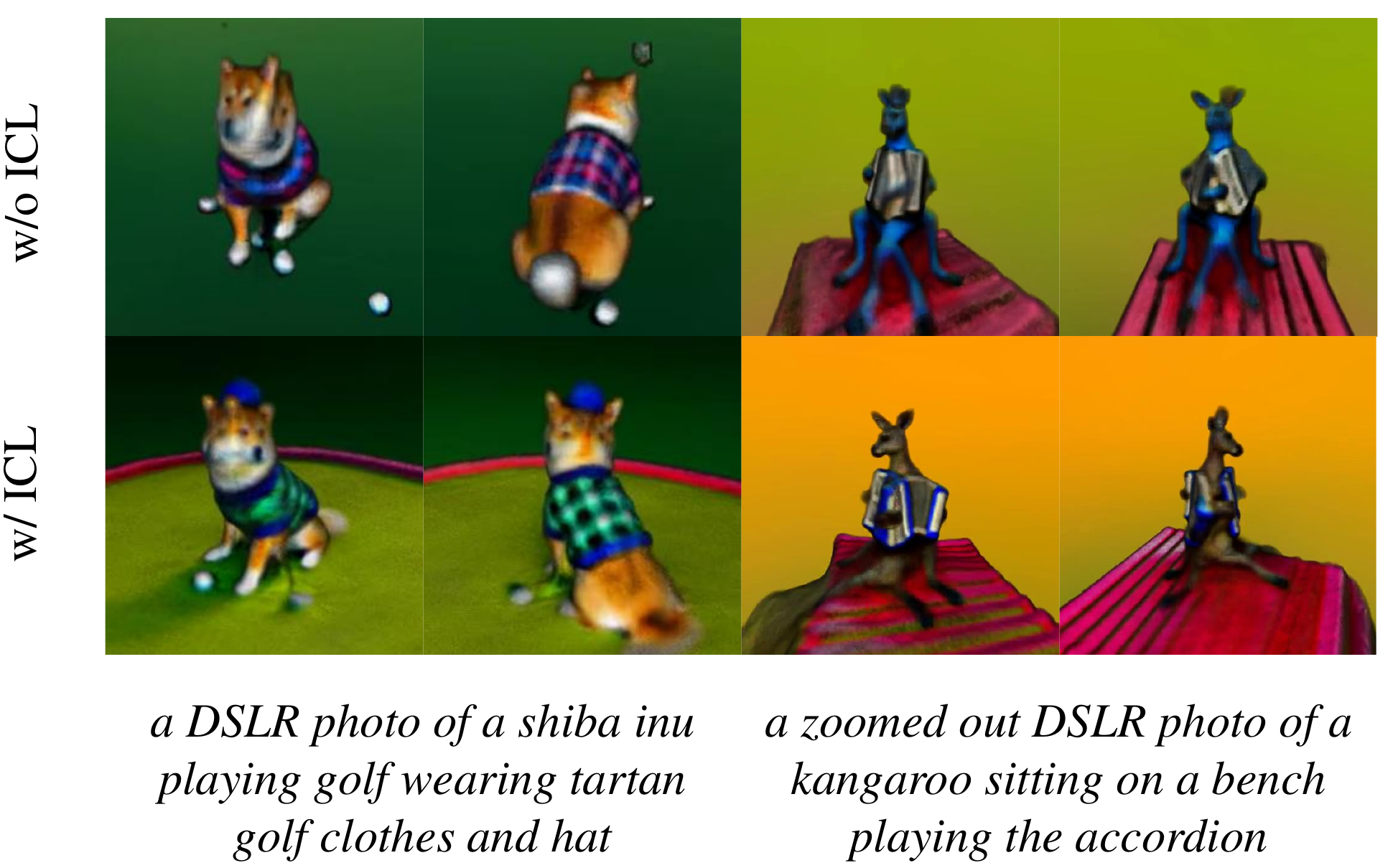}
        \caption{Text-to-3D failure cases.}
        \label{fig:fail_3d}
    \end{minipage}
    
\end{figure}

\textbf{Video-to-4D}
Since the video-to-4D task in this paper is very challenging, our method actually has many failure cases. For example, we fail to generate the dynamic object when the motion is complex or abrupt, as shown in Figure~\ref{fig:fail_4d}. In Figure~\ref{fig:fail_4d}, the dog's tail disappears in the second image because the tail is occluded in the input image when $t=t_2$. The frog, which is jumping up and down fast, gets blurry when $t=t_1$.

\textbf{Text-to-3D}
When applying ICL in text-to-3D, we find some multi-face cases that could not be alleviated, and we show them in Figure~\ref{fig:fail_3d}.

\end{document}